\documentclass{article}

\usepackage[preprint]{neurips_2026}

\usepackage{amsmath}
\usepackage[utf8]{inputenc} 
\usepackage[T1]{fontenc}    
\usepackage{hyperref}       
\usepackage{url}            
\usepackage{booktabs}       
\usepackage{amsfonts}       
\usepackage{nicefrac}       
\usepackage{microtype}      
\usepackage{xcolor}         
\usepackage[pdftex]{graphicx} 

\title{HISPO: Hierarchical Importance-Sampling Policy Optimization with Entropy-Derived Segments}

\author{
  Quoc-Vinh Lai-Dang\\
  Cho Chun Shik Graduate School of Mobility\\
  KAIST\\
  \texttt{ldqvinh@kaist.ac.kr} \\
  \And
  Hyo-Sang Shin\thanks{Corresponding author.}\\
  Cho Chun Shik Graduate School of Mobility\\
  KAIST\\
  \texttt{hyosangshin@kaist.ac.kr}
}

\begin{document}

\maketitle

\begin{abstract}
Reinforcement learning with verifiable rewards (RLVR) has become a central
approach for improving mathematical reasoning in language models, but
long-form completions introduce a difficult credit-assignment problem: different
parts of a solution trace may contribute unevenly to final correctness.
Existing policy-optimization objectives for RLVR commonly apply importance-sampling
correction at either the token level (GRPO, DAPO) or the sequence level (GSPO), imposing different
granularities for assigning credit across a response. 
We introduce \textit{\textbf{H}ierarchical \textbf{I}mportance-\textbf{S}ampling \textbf{P}olicy \textbf{O}ptimization} (\textbf{HISPO}), a segment-level policy-optimization method that
constructs rollout-time entropy-derived contiguous segments, assigns soft
entropy-based saliency weights, and applies clipped importance-sampling
correction at the segment granularity. This provides an intermediate correction unit between
token-level GRPO/DAPO and sequence-level GSPO. 
We evaluate HISPO by fine-tuning
Qwen3-1.7B-Base on mathematical reasoning tasks. Across six benchmarks, HISPO
improves Pass@8 over the strongest baseline on all benchmarks and matches or
exceeds the strongest baseline in Acc@8 on five of them. On AIME25, HISPO
improves over GRPO by $+3.75$ Acc@8 and $+3.78$ Pass@8, and over GSPO by
$+2.50$ Acc@8 and $+1.27$ Pass@8. These results suggest that segment-level
correction is a promising granularity for RLVR in long-form mathematical
reasoning.
\end{abstract}

\section{Introduction}
By 2026, frontier language models increasingly expose reasoning effort as an
explicit inference-time capability. Systems such as GPT-5.5 reasoning models,
Claude 4.x with adaptive or extended thinking, Gemini 3.1 Pro and Deep Think,
Qwen3, and DeepSeek-R1 illustrate a broader shift toward spending more
computation on difficult mathematical, scientific, coding, and agentic tasks
\citep{openai2026gpt55,anthropic2026extendedthinking,google2026gemini31pro,
google2026geminideepthink,yang2025qwen3,deepseekai2025deepseekr1}. This trend
makes reinforcement learning with verifiable rewards (RLVR) especially
attractive for mathematical reasoning, since final answers can often be checked
automatically. At the same time, long completions create a difficult
credit-assignment problem: different tokens, steps, or reasoning regions may
contribute unevenly to final correctness.

Modern RLVR methods inherit this issue from PPO-style clipped policy
optimization~\citep{schulman2017ppo}. Token-level methods such as GRPO and
DAPO provide fine-grained correction, but can be sensitive to noisy local ratios
and length-dependent aggregation~\citep{shao2024deepseekmath,yu2025dapo}.
Sequence-level methods such as GSPO improve stability by correcting the whole
response, but may treat heterogeneous reasoning traces as a single update unit
\citep{zheng2025gspo}. Recent work on length-unbiased sequence optimization
further highlights that response length can interact nontrivially with
sequence-level objectives \citep{liu2026lengthunbiased}. These limitations
motivate an intermediate correction granularity.

Several related directions also emphasize intermediate reasoning structure.
Process-supervision methods use step-level signals, but often require
intermediate labels or process reward models
\citep{lightman2023letsverify,wang2024mathshepherd}. Concurrent work on SPO
studies segment-level advantage estimation, while DHPO mixes token- and
sequence-level importance ratios, including entropy-guided variants
\citep{guo2025spo,min2026dhpo}. Our approach differs by using entropy to form
contiguous reasoning segments and applying clipped importance-sampling
correction directly at the segment level.

We introduce \textit{\textbf{H}ierarchical \textbf{I}mportance-\textbf{S}ampling \textbf{P}olicy \textbf{O}ptimization} (\textbf{HISPO}), a segment-level policy-optimization objective for RLVR in
long-form mathematical reasoning. HISPO partitions each response into
entropy-derived contiguous segments, applies non-maximum suppression and
segment-level weighting to construct stable correction units, and performs
clipped importance-sampling correction at the segment granularity. This creates
a middle ground between token-level GRPO/DAPO-style correction and
sequence-level GSPO-style correction, without requiring step labels, a process
reward model, a critic, or hybrid token/sequence mixing.

We evaluate HISPO by fine-tuning Qwen3-1.7B-Base \citep{yang2025qwen3} with LoRA \citep{hu2021lora} and comparing against
GRPO, DAPO, and GSPO in a mathematical RLVR platform \citep{sheng2024hybridflow}. In
the selected-checkpoint evaluation across six mathematical
reasoning benchmarks, HISPO improves pass rate (Pass@8) over the strongest baseline on all
benchmarks and matches or exceeds the strongest baseline in accuracy (Acc@8) on five of
six. Due to computational constraints, we use monitoring curves with 4 samples (Acc@4 and Pass@4) for checkpoint selection, diagnostics, and
segment-boundary sensitivity analysis, showing that HISPO's gains are most
visible in multi-sample success rates.

Our contributions are threefold. 
First, we propose entropy-based segment
construction with non-maximum suppression and soft segment-level weighting,
forming intermediate correction units without step-level supervision.
Second, we formulate HISPO, a segment-level
clipped importance-sampling objective for RLVR in long-form mathematical
reasoning. Third, we provide a controlled comparison
against token-level and sequence-level policy-optimization variants, together
with monitoring curves, training diagnostics, selected-checkpoint
evaluation, and segment-boundary sensitivity analysis.

\section{Preliminaries}
\label{sec:preliminaries}

We study RLVR for long-form mathematical reasoning. Given a prompt
$x$, a policy $\pi_\theta$ generates a response
$y=(y_1,\ldots,y_T)$, with token probability
$\pi_\theta(y_t \mid x,y_{<t})$. A verifier provides a binary response-level
reward $R\in\{0,1\}$, and policy-gradient methods update $\theta$ through
score-function terms of the form
$\nabla_\theta \log \pi_\theta(y_t \mid x,y_{<t})$.

For a response $y_i=(y_{i,1},\ldots,y_{i,T_i})$ sampled from the behavior
policy $\pi_{\theta_{\mathrm{old}}}$, the token-level importance ratio is
\begin{equation}
  \rho_{i,t}(\theta)
  =
  \frac{\pi_\theta(y_{i,t} \mid x,y_{i,<t})}
       {\pi_{\theta_{\mathrm{old}}}(y_{i,t} \mid x,y_{i,<t})}.
\end{equation}
Policy-optimization methods can then be compared by their advantage estimates,
normalization schemes, and importance-correction granularities.

\subsection{From PPO to GRPO}

Proximal Policy Optimization (PPO)~\citep{schulman2017ppo} controls policy
updates by clipping the importance ratio between the current and behavior
policies. Omitting auxiliary KL terms for brevity, its clipped surrogate for a
single response is
\begin{equation}
  \mathcal{J}_{\mathrm{PPO}}(\theta)
  \approx
  \frac{1}{T}\sum_{t=1}^{T}
  \min\!\left(
    \rho_t(\theta)\hat A_t,\,
    \mathrm{clip}(\rho_t(\theta),1-\epsilon,1+\epsilon)\hat A_t
  \right),
\end{equation}
where $\hat A_t$ is typically estimated from a learned value function. In LLM
RLVR, however, rewards are usually assigned to complete responses, making token-level
value estimation costly and nontrivial.

Group Relative Policy Optimization (GRPO)~\citep{shao2024deepseekmath} removes
the learned value model by sampling a group of responses
$\{y_j\}_{j=1}^{G}$ for the same prompt and normalizing rewards within the
group. For response $y_i$, the group-relative advantage is
\begin{equation}
  \hat A_i =
  \frac{R(x,y_i)-\mathrm{mean}(\{R(x,y_j)\}_{j=1}^G)}
       {\mathrm{std}(\{R(x,y_j)\}_{j=1}^G)} .
\end{equation}
Omitting KL regularization for readability, the GRPO surrogate is
\begin{equation}
  \mathcal{J}_{\mathrm{GRPO}}(\theta)
  \approx
  \frac{1}{G}\sum_{i=1}^{G}\frac{1}{T_i}
  \sum_{t=1}^{T_i}
  \min\!\left(
    \rho_{i,t}(\theta)\hat A_i,\,
    \mathrm{clip}(\rho_{i,t}(\theta),1-\epsilon,1+\epsilon)\hat A_i
  \right).
\end{equation}
For the active unclipped branch, the gradient contribution is
\begin{equation}
\label{eq:grpo-gradient}
  \nabla_{\theta} \mathcal{J}_{\mathrm{GRPO}}(\theta)
  \approx
  \frac{1}{G}\sum_{i=1}^{G}
  \hat A_i
  \frac{1}{T_i}
  \sum_{t=1}^{T_i}
  \rho_{i,t}(\theta)
  \nabla_\theta
  \log \pi_\theta(y_{i,t} \mid x,y_{i,<t}) .
\end{equation}
Thus, PPO and GRPO share clipped token-level importance correction, but differ
in the source of the advantage signal: PPO uses critic-based token advantages,
whereas GRPO assigns each response a group-normalized outcome advantage. GRPO
therefore avoids a separate value model while retaining token-level likelihood
ratios and per-response length normalization.

\subsection{Token-level policy-gradient loss in DAPO}

Decouple Clip and Dynamic sAmpling Policy Optimization (DAPO)~\citep{yu2025dapo} highlights a reduction issue in GRPO's
per-response averaging: each response receives equal total weight, so tokens in
longer responses receive smaller per-token weight. Its token-level
policy-gradient loss instead normalizes by the total number of generated tokens
while retaining the clipped token-level surrogate:
\begin{equation}
  \mathcal{J}_{\mathrm{DAPO}}(\theta)
  \approx
  \frac{1}{\sum_{j=1}^{G}T_j}
  \sum_{i=1}^{G}\sum_{t=1}^{T_i}
  \min\!\left(
    \rho_{i,t}(\theta)\hat A_i,\,
    \mathrm{clip}(\rho_{i,t}(\theta),1-\epsilon_\text{low},1+\epsilon_\text{high})\hat A_i
  \right).
\end{equation}

For the active unclipped branch, the corresponding gradient is
\begin{equation}
\label{eq:dapo-gradient}
  \nabla_{\theta} \mathcal{J}_{\mathrm{DAPO}}(\theta)
  \approx
  \frac{1}{\sum_{j=1}^{G}T_j}
  \sum_{i=1}^{G}
  \hat A_i
  \sum_{t=1}^{T_i}
  \rho_{i,t}(\theta)
  \nabla_\theta
  \log \pi_\theta(y_{i,t} \mid x,y_{i,<t}) .
\end{equation}

Compared with GRPO's per-response normalization, DAPO gives longer-than-average
responses more total gradient mass and shorter-than-average responses less.
Thus, DAPO mitigates length dilution in the aggregation scheme, but its
correction unit remains the individual token. Other DAPO contributions are
not the focus of the granularity analysis considered here.

\subsection{Sequence-level policy optimization in GSPO}

Group Sequence Policy Optimization (GSPO)~\citep{zheng2025gspo} moves importance correction from tokens to the full
response, motivated by the instability of token-wise ratios and clipping in
long completions. It defines a sequence-level ratio as the geometric mean of
token ratios,
\begin{equation}
  s_i(\theta)
  =
  \left(\prod_{t=1}^{T_i}\rho_{i,t}(\theta)\right)^{1/T_i},
\end{equation}
and applies clipping at the response level:
\begin{equation}
  \mathcal{J}_{\mathrm{GSPO}}(\theta)
  \approx
  \frac{1}{G}\sum_{i=1}^{G}
  \min\!\left(
    s_i(\theta)\hat A_i,\,
    \mathrm{clip}(s_i(\theta),1-\epsilon,1+\epsilon)\hat A_i
  \right).
\end{equation}
For the active unclipped branch, the gradient becomes
\begin{equation}
\label{eq:gspo-gradient}
  \nabla_{\theta} \mathcal{J}_{\mathrm{GSPO}}(\theta)
  \approx
  \frac{1}{G}\sum_{i=1}^{G}
  s_i(\theta)\hat A_i
  \frac{1}{T_i}
  \sum_{t=1}^{T_i}
  \nabla_\theta
  \log \pi_\theta(y_{i,t} \mid x,y_{i,<t}) .
\end{equation}
Thus, unlike GRPO and DAPO, which weight each token gradient by its own
$\rho_{i,t}(\theta)$, GSPO assigns all tokens in a response the same
sequence-level correction factor $s_i(\theta)$. This improves response-level
stability but treats the entire reasoning trace as a single optimization unit,
without distinguishing intermediate segments.

\subsection{Motivation}

The preceding objectives expose a correction-granularity tradeoff. PPO provides
token-level clipping but requires a value model; GRPO removes the value model
through group-relative advantages while retaining token-level ratios and
per-response averaging; DAPO changes the normalization to reduce length dilution
but still updates at the token level; and GSPO moves correction to the whole
response, improving stability but treating the entire reasoning trace as one
unit.

Long-form reasoning responses often contain distinct contiguous phases, such as
problem setup, derivation, revision, and answer formation. This motivates
\textit{\textbf{H}ierarchical \textbf{I}mportance-\textbf{S}ampling \textbf{P}olicy \textbf{O}ptimization} (\textbf{HISPO}),
which introduces segment-level correction as an intermediate granularity between
token-level and sequence-level objectives. The next section defines the segment
construction procedure and derives the corresponding HISPO objective and
gradient structure.

\section{Algorithm}
\label{sec:method}

\subsection{Entropy-based segment detection and weight construction}
\label{subsec:entropy-based-segment-detection-and-weight-construction}

We use token entropy as a detached uncertainty signal along each completion. For
token position $t$ in response $y_i$, entropy is computed once at rollout time
under the behavior policy:
\begin{equation}
  H_{i,t}
  =
  -\sum_{v\in\mathcal{V}} p_{i,t}(v)\log p_{i,t}(v),
  \qquad
  p_{i,t}(v)
  =
  \mathrm{sg}\!\left[
  \pi_{\theta_{\mathrm{old}}}(v\mid x,y_{i,<t})
  \right],
\end{equation}
where $\mathcal{V}$ is the vocabulary and $\mathrm{sg}[\cdot]$ denotes
stop-gradient. The entropy trace, segment boundaries, and segment weights are
therefore fixed during subsequent policy-optimization updates. To reduce local
lexical noise, we smooth the trace with a forward exponential moving average (EMA),
\begin{equation}
  \tilde H_{i,1}=H_{i,1},
  \qquad
  \tilde H_{i,t}
  =
  \alpha H_{i,t} + (1-\alpha)\tilde H_{i,t-1},
  \quad t=2,\ldots,T_i .
\end{equation}
As a one-sided smoother, the EMA can slightly lag abrupt entropy changes, so we
use $\tilde H_i$ as a coarse boundary signal rather than a token-exact semantic
boundary estimate.

Let $q_H\in[0,1]$ be the entropy percentile threshold and $\Delta$ the NMS
window. For each response, we define the response-local cutoff
\begin{equation}
  \tau_{i,H}
  =
  \mathrm{Quantile}_{q_H}
  \left(\{\tilde H_{i,t}\}_{t=2}^{T_i-1}\right),
\end{equation}
and collect high-entropy local maxima,
\begin{equation}
  \mathcal{C}_i
  =
  \left\{
  t \in \{2,\ldots,T_i-1\} :
  \tilde H_{i,t} \ge \tau_{i,H},\;
  \tilde H_{i,t} \ge \tilde H_{i,t-1},\;
  \tilde H_{i,t} > \tilde H_{i,t+1}
  \right\}.
\end{equation}
Internal boundaries are selected by token-order non-maximum suppression (NMS),
\begin{equation}
  \mathcal{B}_i
  =
  \mathrm{NMS}_{\Delta}(\mathcal{C}_i;\tilde H_i),
\end{equation}
Here $\mathrm{NMS}_{\Delta}$ scans candidates from left to right, retains a
candidate $t$ only if $\tilde H_{i,t}$ attains the maximum in a centered
$\Delta$-token neighborhood, breaks ties by scan order, and suppresses later
candidates within $\Delta$ tokens of $t$.
The hyperparameters $(\alpha,q_H,\Delta)$ control smoothing,
percentile-based peak selection, and boundary spacing.

Let the sorted selected boundaries and endpoints be
\begin{equation}
  0=b_{i,0}<b_{i,1}<\cdots<b_{i,K_i}=T_i .
\end{equation}
They define contiguous, non-overlapping segments
\begin{equation}
  \mathcal{S}_i
  =
  \{S_{i,k}=\{b_{i,k-1}+1,\ldots,b_{i,k}\}\}_{k=1}^{K_i}.
\end{equation}
If no internal boundary is selected, we set $K_i=1$ and treat the full response
as one segment.

Finally, each segment receives a softmax-normalized entropy saliency weight:
\begin{equation}
  \bar H_{i,k}
  =
  \frac{1}{|S_{i,k}|}\sum_{t\in S_{i,k}} H_{i,t},
  \qquad
  w_{i,k}
  =
  \frac{\exp(\bar H_{i,k})}
       {\sum_{\ell=1}^{K_i}\exp(\bar H_{i,\ell})}.
\end{equation}
Thus, $\sum_{k=1}^{K_i} w_{i,k}=1$. This unit-temperature softmax softly
distributes saliency mass across segments: higher-entropy segments receive
larger saliency mass, while all segments retain nonzero weight. Here
$|S_{i,k}|$ denotes the length of segment $S_{i,k}$.

\subsection{Hierarchical Importance-Sampling Policy Optimization}

We propose 
\textit{\textbf{H}ierarchical \textbf{I}mportance-\textbf{S}ampling \textbf{P}olicy \textbf{O}ptimization} (\textbf{HISPO}), which applies clipped importance correction at the segment
rather than token or sequence level. For response $y_i$, we first define the
group-normalized response advantage
\begin{equation}
  \hat A_i =
  \frac{R(x,y_i)-\mathrm{mean}(\{R(x,y_j)\}_{j=1}^G)}
       {\mathrm{std}(\{R(x,y_j)\}_{j=1}^G)} .
\end{equation}
Each segment then receives a saliency-weighted advantage
\begin{equation}
  \hat A_{i,k}
  =
  \frac{w_{i,k}}{|S_{i,k}|}\hat A_i ,
\end{equation}
where $w_{i,k}$ is the entropy-based segment weight from the previous
Section~\ref{subsec:entropy-based-segment-detection-and-weight-construction} and $|S_{i,k}|$ is the segment length.

For segment $S_{i,k}$, HISPO defines the segment-level importance ratio as the
geometric mean of token ratios within the segment:
\begin{equation}
  \mu_{i,k}(\theta)
  =
  \exp\!\left(
  \frac{1}{|S_{i,k}|}
  \sum_{t\in S_{i,k}}
  \log \rho_{i,t}(\theta)
  \right)
  =
  \left(
  \prod_{t\in S_{i,k}}\rho_{i,t}(\theta)
  \right)^{1/|S_{i,k}|}.
\end{equation}
The HISPO clipped surrogate is
\begin{equation}
  \mathcal{J}_{\mathrm{HISPO}}(\theta)
  \approx
  \frac{1}{G}\sum_{i=1}^{G}
  \sum_{k=1}^{K_i}
  |S_{i,k}|
  \min\!\left(
    \mu_{i,k}(\theta)\hat A_{i,k},\,
    \mathrm{clip}(\mu_{i,k}(\theta),1-\epsilon,1+\epsilon)\hat A_{i,k}
  \right).
\end{equation}
Equivalently, since $|S_{i,k}|\hat A_{i,k}=w_{i,k}\hat A_i$, HISPO assigns
each response-level advantage across entropy-weighted segments while using
$\mu_{i,k}(\theta)$ as the clipped correction unit. This gives an intermediate
correction granularity between token-level GRPO/DAPO and sequence-level GSPO.

\subsection{Gradient Analysis}

\begin{table}[t]
  \caption{Token-gradient coefficients under a shared group-relative advantage.
The comparison isolates correction granularity: GRPO and DAPO use token-level
ratios, GSPO uses one sequence-level ratio, and HISPO uses entropy-weighted
segment-level ratios.}
  \label{tab:method-gradient-comparison}
  \centering
  \small
  \begin{tabular}{p{0.10\linewidth}p{0.15\linewidth}p{0.22\linewidth}p{0.40\linewidth}}
    \toprule
    Method & Correction unit & Normalization view & Token-gradient coefficients \\
    \midrule
    GRPO & token & per response, then group & $\hat A_i\rho_{i,t}(\theta)/(G T_i) $ \\
    DAPO & token & all valid response tokens & $\hat A_i\rho_{i,t}(\theta)/\sum_j^G T_j$ \\
    GSPO & sequence & response-level correction with token mean & $\hat A_i\left(\prod_{t=1}^{T_i}\rho_{i,t}(\theta)\right)^{1/T_i}/(G T_i)$ \\
    HISPO & segment & saliency segment mass, uniform token share & $\hat A_i \left(\prod_{t\in S_{i,k}}\rho_{i,t}(\theta)\right)^{1/|S_{i,k}|}w_{i,k}/(G |S_{i,k}|)$ \\
    \bottomrule
  \end{tabular}
\end{table}

We compare update coefficients under the active unclipped branch, without
auxiliary KL terms, and with the shared group-relative advantage $\hat A_i$.
Segment boundaries and entropy-derived weights are fixed as described in
Subsection~\ref{subsec:entropy-based-segment-detection-and-weight-construction}.
From the HISPO objective, the gradient contribution is
\begin{equation}
\label{eq:hispo-gradient}
  \nabla_{\theta} \mathcal{J}_{\mathrm{HISPO}}(\theta)
  \approx
  \frac{1}{G}\sum_{i=1}^{G}
  \sum_{k=1}^{K_i}
  \mu_{i,k}(\theta)\hat A_i
  \frac{w_{i,k}}{|S_{i,k}|}
  \sum_{t\in S_{i,k}}
  \nabla_\theta
  \log \pi_\theta(y_{i,t}\mid x,y_{i,<t}) ,
\end{equation}
where
\begin{equation}
  \mu_{i,k}(\theta)
  =
  \left(\prod_{t\in S_{i,k}}\rho_{i,t}(\theta)\right)^{1/|S_{i,k}|}.
\end{equation}
Thus, all tokens in a segment share the same correction ratio
$\mu_{i,k}(\theta)$, while the segment saliency mass $w_{i,k}\hat A_i$ is
distributed uniformly over the segment tokens.

Table~\ref{tab:method-gradient-comparison} summarizes the resulting
token-gradient coefficients, up to common optimizer and batch-scale constants.
For HISPO, the coefficient applies to tokens $t\in S_{i,k}$.

This comparison isolates the role of correction granularity. GRPO and DAPO use
token-specific ratios, GSPO assigns one ratio to the full response, and HISPO
assigns one ratio to each entropy-derived segment.

\section{Experiments}
\label{sec:experiments}

We evaluate HISPO in mathematical RLVR. We first describe the shared training
setup and evaluation protocol, then analyze validation curves, training
diagnostics, selected-checkpoint performance, and hyperparameter sensitivity.

\subsection{Experimental Settings}
\label{subsec:experimental-settings}
All experiments use Qwen3-1.7B-Base~\citep{yang2025qwen3}, a practical scale
for controlled long-reasoning experiments under our compute budget. We train on
the MATH dataset~\citep{hendrycks2021measuring}, which contains 7,500
competition-style mathematical reasoning problems, and evaluate selected
checkpoints on six benchmarks: MATH500, Minerva~\citep{lewkowycz2022solving}, AMC23, AIME24, AIME25, and
OlympiadBench~\citep{he2024olympiadbench}. 
We report accuracy (Acc@$k$) and pass rate (Pass@$k$); Acc@$k$ averages
correctness over the $k$ sampled responses, while Pass@$k$ computes a bootstrap
estimate of best-of-$k$ correctness for each problem and then averages these
per-problem estimates. Formal definitions are given in
Appendix~\ref{app:evaluation-metrics}.

We compare the correction granularities introduced in
Sections~\ref{sec:preliminaries} and~\ref{sec:method}: GRPO
\citep{shao2024deepseekmath} and DAPO~\citep{yu2025dapo} as token-level
methods, GSPO~\citep{zheng2025gspo} as a sequence-level method, and HISPO as
the proposed segment-level method. All methods use the same \texttt{verl}-based
platform~\citep{sheng2024hybridflow} and recipe family. To fit the
available hardware budget, all runs use LoRA fine-tuning~\citep{hu2021lora} with batch size of 32 and mini-batch size of 8. Implementation details are provided in the Appendix. Scaling to larger models, larger batches, longer schedules, and full fine-tuning remains future work.

\subsection{Main Results and Analysis}
\label{subsec:main-results-analysis}

\begin{figure}[t]
  \centering
  \includegraphics[width=\linewidth]{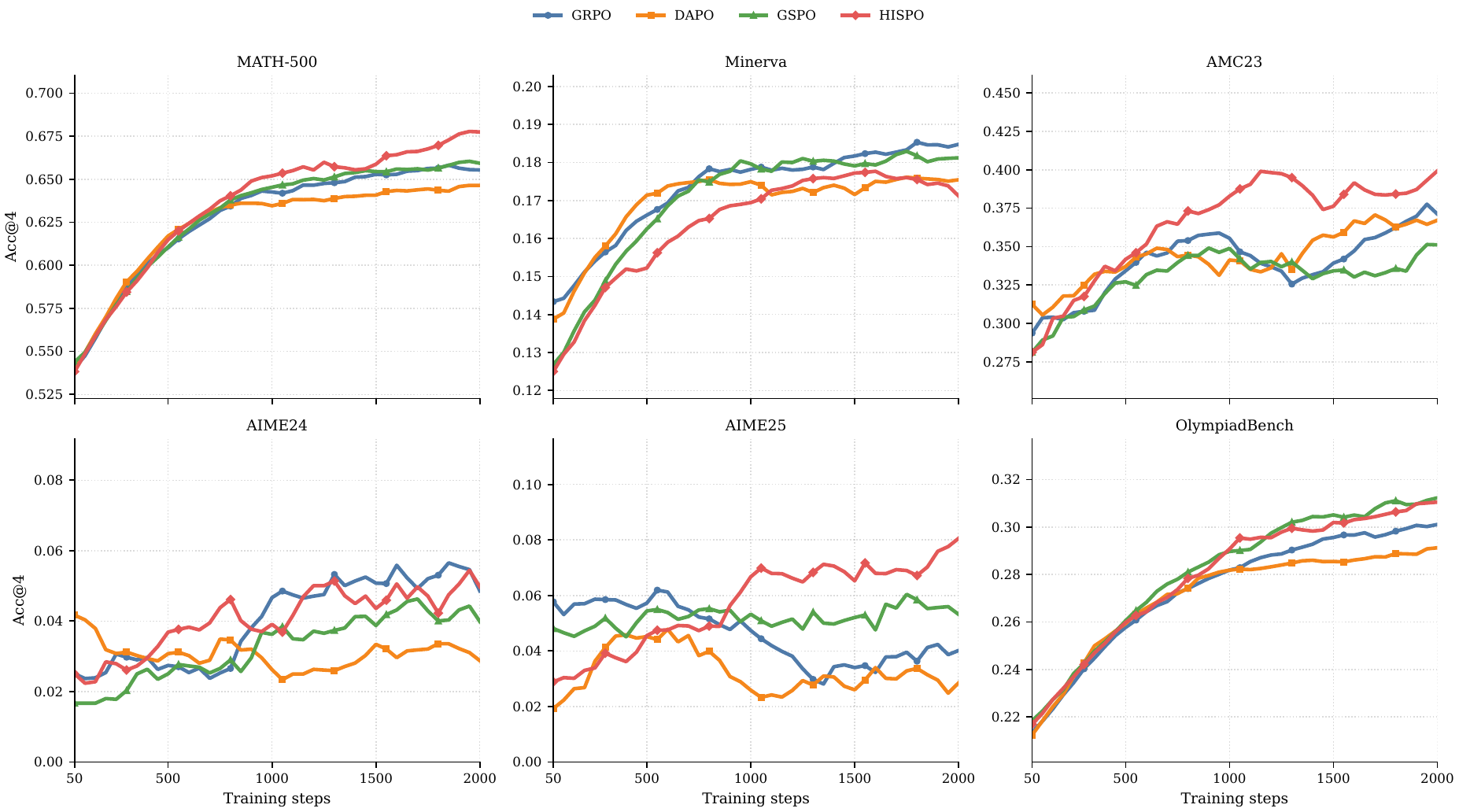}
  \caption{Development checkpoint curves for Acc@4 on the six-benchmark validation suite. HISPO shows steady improvements on most benchmarks, with the clearest gains on AMC23 and AIME25. All curves are smoothed for visual clarity.}
  \label{fig:acc4-postfix-monitoring}
\end{figure}

\begin{figure}[t]
  \centering
  \includegraphics[width=\linewidth]{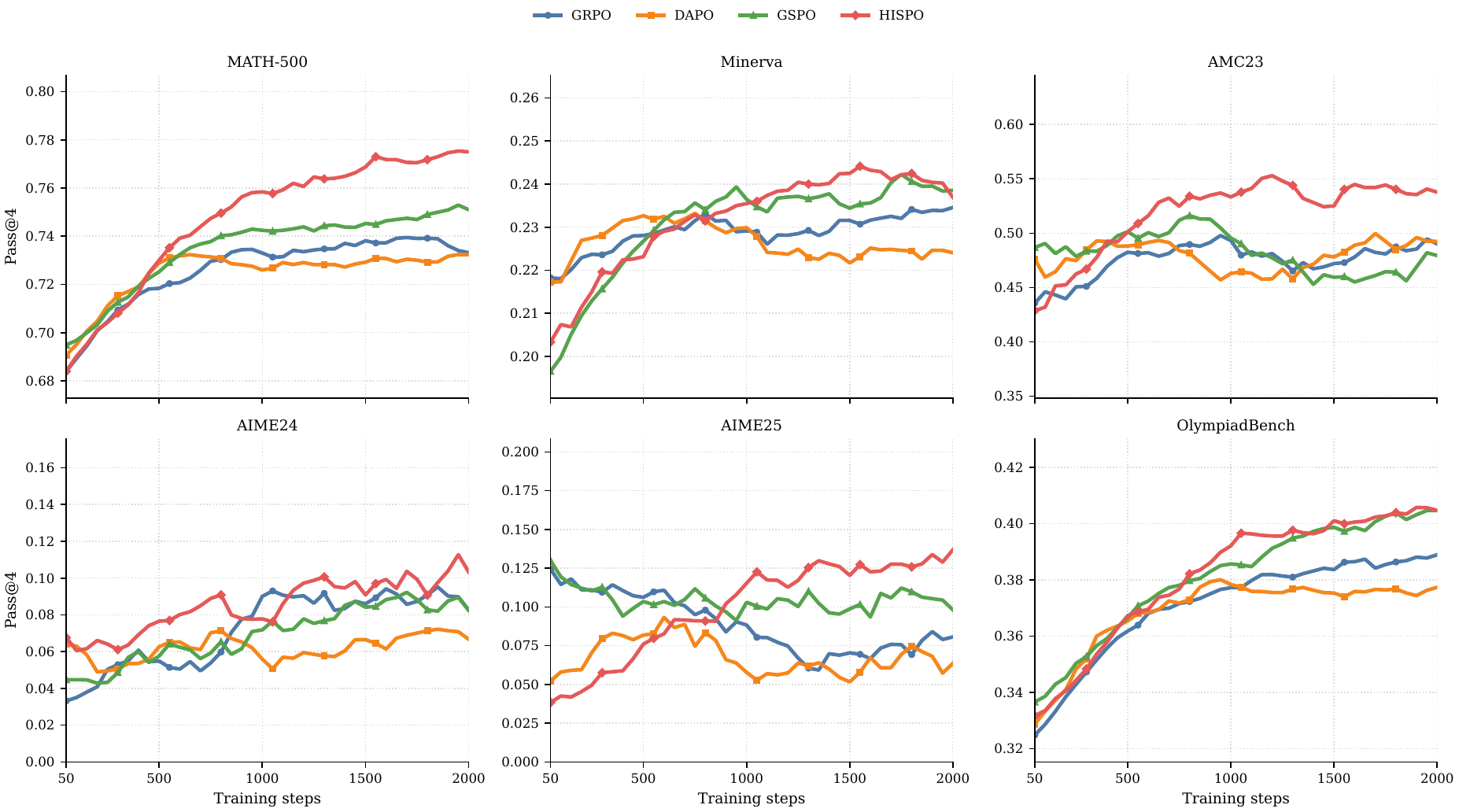}
  \caption{Development checkpoint curves for Pass@4 on the six-benchmark validation suite. HISPO provides the most consistent multi-sample gains, especially on AMC23 and AIME25. All curves are smoothed for visual clarity.}
  \label{fig:pass4-postfix-monitoring}
\end{figure}

For validation-curve monitoring, we generate four responses per problem and
track Acc@4 and Pass@4 across training
(Figures~\ref{fig:acc4-postfix-monitoring} and
\ref{fig:pass4-postfix-monitoring}). In the displayed trajectories, HISPO shows
broad improvements, ending at approximately $0.68/0.78$ Acc@4/Pass@4 on
MATH500, $0.40/0.54$ on AMC23, $0.08/0.13$ on AIME25, and $0.31/0.41$ on
OlympiadBench, matching or exceeding the strongest baseline on most of these
benchmarks. The clearest gains appear on AMC23 and AIME25, where several
baselines plateau or regress while HISPO continues to improve, suggesting better
multi-sample success as well as higher average correctness. Minerva is the main
exception: GRPO and GSPO retain a small Acc@4 advantage, although HISPO remains
close in Pass@4. These monitoring trends motivate a shared checkpoint-selection
protocol for the final comparison.

\begin{figure}[t]
  \centering
  \includegraphics[width=\linewidth]{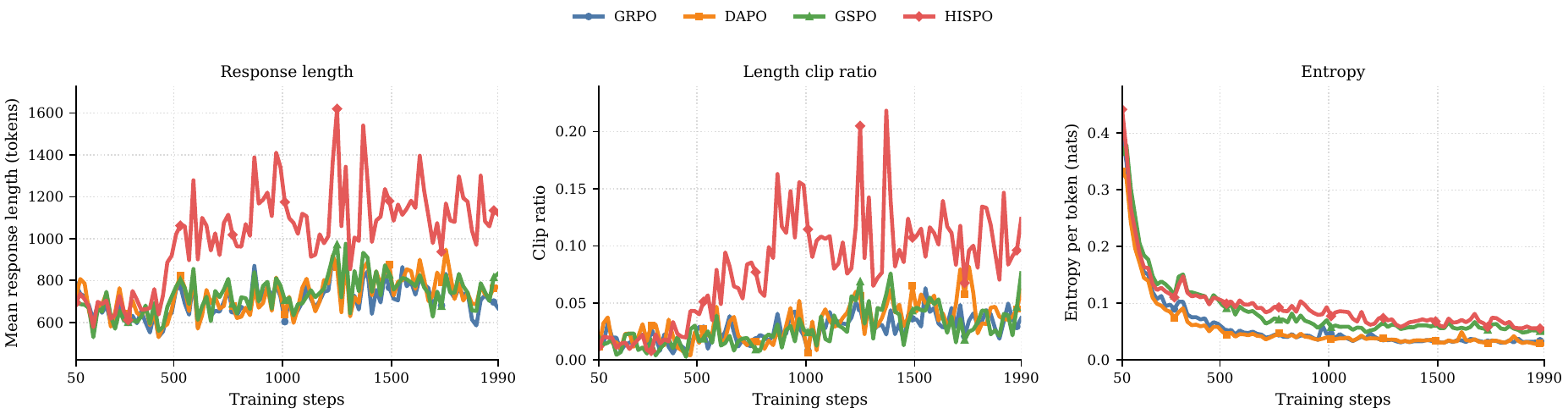}
  \caption{Training diagnostics for response length, length clipping, and
  entropy. HISPO produces longer, higher-entropy generations, indicating
  stronger length pressure and more sustained exploration during training. All curves are smoothed for visual clarity.}
  \label{fig:training-diagnostics}
\end{figure}

Figure~\ref{fig:training-diagnostics} summarizes response length,
length-clipping ratio, and entropy during training. HISPO follows a distinct
regime: its mean response length rises from roughly $650$ to above $1{,}100$
tokens, with peaks near $1{,}600$, while the baselines mostly remain in the
$600$--$1000$ range; its clipping ratio also spikes near $0.20$, indicating
stronger length pressure. At the same time, HISPO maintains higher entropy than
DAPO and remains comparable to or above GRPO and GSPO after the initial drop.
These diagnostics contextualize the validation gains, but remain correlational:
they do not by themselves prove that entropy-induced boundaries are semantic
boundaries or that segmentation is the causal source of the improvements.

\begin{table}[t]
  \caption{Selected-checkpoint top-1 @8 comparison on six benchmarks.
  \textbf{Bold} and \underline{underlined} entries mark the best
  and second-best distinct values within each column, with ties all bolded.
  The final row reports HISPO minus the largest non-HISPO value in percentage
  points. HISPO improves Pass@8
  on all six benchmarks and matches or exceeds the best Acc@8 baseline on five
  of six.}
  \label{tab:main-selected-checkpoint-at8}
  \centering
  \scriptsize
  \setlength{\tabcolsep}{2.0pt}
  \resizebox{\linewidth}{!}{%
  \begin{tabular}{lrrrrrrrrrrrr}
    \toprule
    & \multicolumn{2}{c}{\textbf{MATH500}} 
    & \multicolumn{2}{c}{\textbf{Minerva}}
    & \multicolumn{2}{c}{\textbf{AMC23}} 
    & \multicolumn{2}{c}{\textbf{AIME24}}
    & \multicolumn{2}{c}{\textbf{AIME25}} 
    & \multicolumn{2}{c}{\textbf{OlympiadBench}} \\
    \cmidrule(lr){2-3} \cmidrule(lr){4-5} \cmidrule(lr){6-7}
    \cmidrule(lr){8-9} \cmidrule(lr){10-11} \cmidrule(lr){12-13}
    \textbf{Method} & Acc@8 & Pass@8 & Acc@8 & Pass@8 & Acc@8 & Pass@8
    & Acc@8 & Pass@8 & Acc@8 & Pass@8 & Acc@8 & Pass@8 \\
    \midrule
    GRPO & 64.70 & 77.33 & \textbf{18.57} & 26.48
    & \underline{37.81} & 57.73 & \textbf{5.42} & 12.08
    & 4.58 & 13.68 & 29.65 & 43.92 \\
    DAPO & 64.75 & 76.71 & 17.28 & 26.32
    & 34.69 & 53.37 & \underline{2.50} & 7.77
    & 5.00 & 13.61 & 29.15 & 44.59 \\
    GSPO & \underline{65.55} & \underline{78.65}
    & \underline{18.34} & \underline{27.62}
    & 36.56 & \underline{60.73} & \textbf{5.42} & \underline{14.19}
    & \underline{5.83} & \underline{16.19}
    & \underline{30.99} & \underline{45.93} \\
    \textbf{HISPO} & \textbf{68.00} & \textbf{81.25} & 17.51 & \textbf{29.11}
    & \textbf{40.00} & \textbf{64.42} & \textbf{5.42} & \textbf{16.09}
    & \textbf{8.33} & \textbf{17.46}
    & \textbf{31.49} & \textbf{46.72} \\
    \midrule
    HISPO $\Delta$ & \textcolor{green!45!black}{\textbf{+2.45}}
    & \textcolor{green!45!black}{\textbf{+2.60}} & \textcolor{red!70!black}{\textbf{-1.06}}
    & \textcolor{green!45!black}{\textbf{+1.49}} & \textcolor{green!45!black}{\textbf{+2.19}}
    & \textcolor{green!45!black}{\textbf{+3.69}} & {\textbf{+0.00}}
    & \textcolor{green!45!black}{\textbf{+1.90}} & \textcolor{green!45!black}{\textbf{+2.50}}
    & \textcolor{green!45!black}{\textbf{+1.27}} & \textcolor{green!45!black}{\textbf{+0.50}}
    & \textcolor{green!45!black}{\textbf{+0.79}} \\
    \bottomrule
  \end{tabular}}
\end{table}

\begin{table}[t]
  \caption{Development-level @4 sensitivity to the entropy-boundary NMS window
($\mathrm{NMS}_{\Delta}$). The Average columns macro-average the three displayed
benchmarks: MATH500, Minerva, and AMC23. \textbf{Bold} and
\underline{underlined} entries mark the best and second-best distinct values
within each metric column, with ties all underlined where they are second-best.
$\mathrm{NMS}_{64}$ obtains the strongest average Acc@4 and Pass@4 among the
tested windows.}
  \label{tab:training-log-at4-sensitivity}
  \centering
  \scriptsize
  \setlength{\tabcolsep}{2.0pt}
  \makebox[\linewidth][c]{%
  \resizebox{0.78\linewidth}{!}{%
  \begin{tabular}{lrrrrrrrr}
    \toprule
    & \multicolumn{2}{c}{\textbf{MATH500}}
    & \multicolumn{2}{c}{\textbf{Minerva}}
    & \multicolumn{2}{c}{\textbf{AMC23}}
    & \multicolumn{2}{c}{\textbf{Average}} \\
    \cmidrule(lr){2-3} \cmidrule(lr){4-5}
    \cmidrule(lr){6-7} \cmidrule(lr){8-9}
    \textbf{Method} & Acc@4 & Pass@4 & Acc@4 & Pass@4
    & Acc@4 & Pass@4 & Acc@4 & Pass@4 \\
    \midrule
    GRPO & 65.62 & 73.24 & \underline{18.66} & 23.71
    & 41.88 & 57.19 & 42.05 & 51.38 \\
    DAPO & 65.52 & 75.14 & 17.74 & 23.79
    & 40.00 & 51.78 & 41.09 & 50.24 \\
    GSPO & \underline{66.33} & \underline{75.33}
    & 18.20 & 23.55 & 40.62 & \underline{57.35}
    & 41.72 & \underline{52.08} \\
    \midrule
    HISPO-$\mathrm{NMS}_{32}$ & 65.50 & 74.77
    & \textbf{18.93} & \underline{24.96} & \underline{43.12} & 55.76
    & \underline{42.52} & 51.83 \\

    HISPO-$\mathrm{NMS}_{64}$ & \textbf{67.09} & \textbf{77.16}
    & 18.57 & \textbf{25.50} & \textbf{45.00} & \textbf{62.01}
    & \textbf{43.55} & \textbf{54.89} \\

    HISPO-$\mathrm{NMS}_{128}$ & 62.35 & 72.04 & 18.01 & 23.45
    & \underline{43.12} & 56.19 & 41.16 & 50.56 \\

    \bottomrule
  \end{tabular}}}
\end{table}

For the main selected-checkpoint comparison, we select one checkpoint per method
by macro-averaging Acc@4 across the six validation benchmarks, then evaluate the
selected checkpoint with eight responses per problem on the same suite. Thus,
Table~\ref{tab:main-selected-checkpoint-at8} is a shared selected-checkpoint
benchmark-suite comparison, not a fully held-out test estimate. Under this
protocol, HISPO improves Pass@8 over the strongest non-HISPO baseline on all six
benchmarks, with an average gain of $+1.96$ points and the largest gains on
AMC23 ($+3.69$), MATH500 ($+2.60$), and AIME24 ($+1.90$). For Acc@8, HISPO
matches or exceeds the strongest non-HISPO baseline on five of six benchmarks,
with Minerva the only negative case ($-1.06$). These trends align with the
monitoring curves and diagnostics, where HISPO's longer, higher-entropy
generations are associated with stronger multi-sample success.
Appendix~\ref{app:uncertainty-diagnostics} reports fresh HISPO--GSPO bootstrap
diagnostics: macro Pass@8 point estimates remain positive, but intervals cross
zero and Minerva/AIME24 show sampling sensitivity.

\subsection{Hyperparameter Sensitivity Analysis}
\label{subsec:hyperparameter-sensitivity}

Table~\ref{tab:training-log-at4-sensitivity} evaluates HISPO's sensitivity to
the NMS window ($\mathrm{NMS}_{\Delta}$) used for entropy-based segment-boundary
selection on development-level @4 metrics over MATH500, Minerva, and AMC23.
The Average columns macro-average these three displayed benchmarks.
$\mathrm{NMS}_{64}$ is strongest overall, improving the best baseline Acc@4
average from $42.05$ (GRPO) to $43.55$ and the best baseline Pass@4 average
from $52.08$ (GSPO) to $54.89$. The largest gains appear on AMC23
($45.00/62.01$, or $+3.12/+4.66$ over the strongest baseline). It also improves
MATH500 from $66.33/75.33$ to $67.09/77.16$, while Minerva remains less uniform:
$\mathrm{NMS}_{32}$ obtains the best Acc@4 ($18.93$), but $\mathrm{NMS}_{64}$
obtains the best Pass@4 ($25.50$). Compared with $\mathrm{NMS}_{32}$
($42.52/51.83$ average) and $\mathrm{NMS}_{128}$ ($41.16/50.56$ average),
$\mathrm{NMS}_{64}$ provides the best average among tested NMS windows,
supporting it as the default configuration for the main selected-checkpoint
experiments.

\section{Limitations and Future Work}
\label{sec:limitations}

HISPO uses entropy-derived segments as intermediate units and entropy-softmax
weights to allocate advantage mass, but our experiments evaluate HISPO as a
combined method and do not establish the mechanism causally. The diagnostics are
correlational: they do not prove that entropy boundaries are semantic reasoning
units, that entropy-softmax weighting independently contributes to the gains, or
that segmentation alone explains the improvements. Our evaluation is limited to
mathematical RLVR with Qwen3-1.7B-Base and LoRA under compute constraints; since
HISPO can increase response length and clipping pressure, its benefits should be
weighed against token-budget and inference-cost constraints.

Future work should test larger-scale and full-finetuning regimes, additional
seeds, and checkpoint-selection protocols that separate model selection from
final reporting. It should also isolate HISPO's components by comparing
entropy-based boundaries with fixed or random alternatives and entropy-softmax
weights with uniform segment weights. Further theory, boundary diagnostics, and
extensions beyond mathematical RLVR remain important directions.

\section{Related Work}

\paragraph{Policy-optimization granularity in RLVR.}
HISPO builds on clipped importance-ratio policy optimization from
PPO~\citep{schulman2017ppo} and its recent adaptations to mathematical
RLVR. GRPO replaces the learned value model with group-relative
advantages~\citep{shao2024deepseekmath}, while DAPO retains token-level
correction but revises the aggregation and normalization for long-CoT
training~\citep{yu2025dapo}. GSPO instead moves correction to the sequence
level with a response-level importance ratio and clipping rule
\citep{zheng2025gspo}. These methods expose a granularity tradeoff: token-level
correction is fine-grained but noisy and length-sensitive, whereas
sequence-level correction is more stable but coarse for heterogeneous reasoning
traces. Recent work explores nearby design points: SPO studies segment-level advantage
estimation~\citep{guo2025spo}, DHPO mixes token- and sequence-level importance
ratios~\citep{min2026dhpo}, and LUSPO corrects length bias in sequence-level
optimization~\citep{liu2026lengthunbiased}. HISPO differs by combining
entropy-derived contiguous segmentation, segment-level geometric-mean importance
correction, and entropy-softmax segment weighting inside a clipped RLVR
objective; Appendix~\ref{app:closest-methods} provides a compact comparison with the
closest adjacent policy-optimization design points, including SPO, DHPO, and
LUSPO.

\paragraph{Entropy and key-token signals in RLVR.}
RLVR is effective for long-form reasoning when final answers are verifiable, as
illustrated by DeepSeek-R1~\citep{deepseekai2025deepseekr1}. Recent studies also
show that entropy and uncertainty reveal important structure in reasoning
updates. \citet{wang2025beyond8020} identify high-entropy minority tokens as
reasoning forks, while \citet{cheng2025reasoningexploration}
link high-entropy regions to pivotal tokens and reflective
behaviors. Related entropy-mechanism work
studies entropy collapse during RL~\citep{cui2025entropy}, and RL-ZVP uses
entropy-guided advantage shaping for zero-variance prompts~\citep{le2025noprompt}.
HISPO differs by using entropy to form contiguous segment-level correction
units, rather than only weighting, masking, or regularizing token-level updates.

\section{Conclusion}

Long-form mathematical RLVR poses a credit-assignment challenge because different
parts of an extended reasoning trace may contribute unevenly to final
correctness. We introduced \textbf{HISPO}, a segment-level policy-optimization
objective that constructs entropy-derived contiguous segments and applies
clipped importance-sampling correction at the segment granularity, providing an
intermediate design point between token-level and sequence-level correction. In
a shared mathematical reasoning evaluation with Qwen3-1.7B-Base, HISPO improves
selected-checkpoint Pass@8 over the strongest baseline on all six benchmarks and
matches or exceeds the strongest baseline in Acc@8 on five of six. Together with
the validation curves, training diagnostics, and segment-boundary sensitivity
analysis, these results suggest that segment-level correction is a promising
granularity for RLVR in long-form reasoning, while further work is needed to
validate the segmentation mechanism, scale the training regime, and extend the
approach beyond verifiable mathematical rewards.

\clearpage
\appendix

\section{Implementation Details}
\label{app:implementation-details}

All methods use the same \texttt{verl} training platform and
Qwen3-1.7B-Base~\citep{yang2025qwen3}. To fit the single-GPU hardware budget,
we train with LoRA~\citep{hu2021lora}, following recent evidence that LoRA can
be competitive in some RL post-training settings
\citep{schulman2025lorawithoutregret}. All runs use one RTX 5090 GPU with 32GB
memory, LoRA rank $r=1$, LoRA scale $\alpha=32$, and all linear modules as
adaptation targets.

During training, each prompt is sampled with eight responses. We monitor
development performance every 50 steps with four samples per prompt. After
2,000 training steps, we sweep checkpoints on the six-benchmark suite, select
one checkpoint per method by macro-averaging Acc@4 across benchmarks, and
re-evaluate the selected checkpoint with eight samples per prompt on MATH500,
Minerva, AMC23, AIME24, AIME25, and OlympiadBench. Shared hyperparameters are
listed in Table~\ref{tab:appendix-shared-hparams}. For HISPO, the default
segment-detector settings are EMA smoothing $\alpha=0.1$, NMS window
$\Delta=64$, and entropy percentile threshold $q_H=0.60$. The implementation
distributes segment saliency mass using a minimum segment-length floor
$L_{\min}=16$, replacing $|S_{i,k}|$ by $\max(|S_{i,k}|, L_{\min})$ in the
per-token mass denominator.

\begin{table}[t]
\caption{Shared training and evaluation hyperparameters.}
\label{tab:appendix-shared-hparams}
\centering
\scriptsize
\begin{tabular}{@{}p{0.34\linewidth}p{0.22\linewidth}@{}}
\toprule
Hyperparameter & Value \\
\midrule
Train batch / mini-batch / micro-batch & 32 / 8 / 2 \\
Training samples per prompt & 8 \\
Monitoring samples per prompt & 4 \\
Prompt / response maximum length & 512 / 4096 \\
Optimizer & AdamW \\
Learning rate / warmup / scheduler & $3\times 10^{-6}$ / 0.05 / constant \\
Train temperature / top-$p$ / top-$k$ & 1.0 / 1.0 / $-1$ \\
Validation temperature / top-$p$ / top-$k$ & 1.0 / 0.7 / $-1$ \\
\bottomrule
\end{tabular}
\end{table}

\subsection{Compute Resources}
\label{app:compute-resources}

All reported experiments use one RTX 5090 GPU with 32GB memory, so wall-clock
hours equal GPU-hours. Table~\ref{tab:appendix-compute} summarizes training,
checkpoint-sweep, and final-evaluation costs. The total reported compute is
approximately 570--645 GPU-hours and excludes preliminary debugging runs.

\begin{table}[t]
\caption{Compute resources for the reported experiments.}
\label{tab:appendix-compute}
\centering
\scriptsize
\begin{tabular}{@{}p{0.38\linewidth}p{0.15\linewidth}p{0.22\linewidth}p{0.15\linewidth}@{}}
\toprule
Experiment group & Number of runs & Wall-clock time & Total GPU-hours \\
\midrule
Baselines: GRPO, DAPO, GSPO & 3 & 75--85 h & 225--255 \\
HISPO variants: $\mathrm{NMS}_{32}, \mathrm{NMS}_{64}, \mathrm{NMS}_{128}$ & 3 & 85--100 h & 255--300 \\
Full six-benchmark checkpoint sweep & 4 & 21 h & 84 \\
Selected-checkpoint (Acc@8, Pass@8) evaluation & 4 & 5.5 h total & 5.5 \\
\midrule
Total reported compute & -- & -- & 570--645 \\
\bottomrule
\end{tabular}
\end{table}

\subsection{Evaluation Metrics}
\label{app:evaluation-metrics}

For problem $i$ and sampled response $m\in\{1,\ldots,k\}$, let
$s_{i,m}\in\{0,1\}$ denote verifier correctness. Acc@$k$ is the benchmark mean
of per-problem sample correctness:
\begin{equation}
  \hat a_i^{(k)}
  =
  \frac{1}{k}\sum_{m=1}^{k} s_{i,m},
  \qquad
  \mathrm{Acc@}k
  =
  \frac{1}{N}\sum_{i=1}^{N}\hat a_i^{(k)} .
\end{equation}

Pass@$k$ follows the evaluator's \texttt{best@k/mean} reducer. For each
problem, the evaluator draws $B$ bootstrap resamples of size $k$ from the $k$
response scores and applies a maximum reducer. With
$I_{i,b,j}\sim\mathrm{Unif}(\{1,\ldots,k\})$, we define
\begin{equation}
  \hat p_i^{(k)}
  =
  \frac{1}{B}\sum_{b=1}^{B}
  \max_{j=1,\ldots,k} s_{i,I_{i,b,j}},
  \qquad
  \mathrm{Pass@}k
  =
  \frac{1}{N}\sum_{i=1}^{N}\hat p_i^{(k)} .
\end{equation}
We use $B=1000$ and seed $42$. Thus, Pass@$k$ is the evaluator's bootstrap
estimate of best-of-$k$ correctness, rather than the literal observed
any-correct indicator.

\subsection{Closest Method Comparison}
\label{app:closest-methods}

Table~\ref{tab:appendix-closest-methods} summarizes the closest adjacent
policy-optimization design points and highlights HISPO's specific combination
of entropy-derived segmentation, segment-level importance-ratio correction, and
entropy-softmax saliency weighting.

\begin{table}[t]
\caption{Closest policy-optimization design points for long-form RLVR. HISPO
differs by combining entropy-derived contiguous segmentation, segment-level
geometric-mean importance-ratio correction, and entropy-softmax saliency
weighting inside a clipped RLVR objective.}
\label{tab:appendix-closest-methods}
\centering
\scriptsize
\setlength{\tabcolsep}{2.5pt}
\begin{tabular}{@{}p{0.18\linewidth}p{0.20\linewidth}p{0.25\linewidth}p{0.28\linewidth}@{}}
\toprule
Method & Main granularity & Main mechanism & Relation to HISPO \\
\midrule
SPO~\citep{guo2025spo}
& segment advantage
& segment-level advantage estimation with Monte Carlo estimation
& Shares segment-level motivation, but focuses on advantage estimation rather
than entropy-derived segment-level importance-ratio correction. \\

DHPO~\citep{min2026dhpo}
& hybrid token/sequence
& mixing token- and sequence-level importance ratios
& Blends token and sequence ratios, whereas HISPO defines contiguous
segment-level correction units. \\

LUSPO~\citep{liu2026lengthunbiased}
& sequence correction
& length-unbiased sequence-level optimization
& Addresses response-length bias in sequence-level correction rather than
segment-level correction within a response. \\

HISPO (Ours)
& entropy-derived segment correction
& entropy-derived contiguous segments, geometric-mean segment ratios, and
entropy-softmax saliency weights
& Applies clipped importance-sampling correction directly at the entropy-derived
segment granularity. \\
\bottomrule
\end{tabular}
\end{table}

\subsection{Selected-Checkpoint Uncertainty Diagnostics}
\label{app:uncertainty-diagnostics}

As a sampling-variability diagnostic, we ran a fresh $k=8$ evaluation of the
selected HISPO and GSPO checkpoints on the six-benchmark suite. This diagnostic
does not vary training seeds, checkpoint-selection rules, the benchmark problem
sets themselves, or model initialization; it estimates only problem-level
uncertainty for a fresh selected-checkpoint rerun. For each benchmark, we compute
paired bootstrap intervals over matched problem IDs using the metrics from
Appendix~\ref{app:evaluation-metrics}. These intervals are diagnostic for the
fresh HISPO--GSPO comparison, not confidence intervals for the exact sampled
generations in Table~\ref{tab:main-selected-checkpoint-at8}.

\begin{table}[t]
\caption{Fresh selected-checkpoint HISPO--GSPO uncertainty diagnostic on the
six-benchmark suite. Values are percentages; intervals are 95\% paired
bootstrap intervals over matched problem IDs.}
\label{tab:app-full-fresh-ci}
\centering
\scriptsize
\setlength{\tabcolsep}{3.0pt}
\begin{tabular}{@{}p{0.20\linewidth}p{0.10\linewidth}rrrp{0.12\linewidth}@{}}
\toprule
Benchmark & Metric & HISPO & GSPO & $\Delta$ & 95\% CI \\
\midrule
MATH500 & Acc@8 & 68.15 & 65.70 & +2.45 & [+0.70, +4.25] \\
MATH500 & Pass@8 & 81.38 & 78.50 & +2.88 & [+0.96, +4.93] \\
Minerva & Acc@8 & 16.59 & 18.80 & -2.21 & [-4.14, -0.41] \\
Minerva & Pass@8 & 27.30 & 27.58 & -0.28 & [-3.02, +2.53] \\
AMC23 & Acc@8 & 41.25 & 39.69 & +1.56 & [-6.56, +10.00] \\
AMC23 & Pass@8 & 66.95 & 63.47 & +3.47 & [-7.98, +15.55] \\
AIME24 & Acc@8 & 3.75 & 6.25 & -2.50 & [-6.67, +0.42] \\
AIME24 & Pass@8 & 10.94 & 14.28 & -3.34 & [-8.95, +0.13] \\
AIME25 & Acc@8 & 7.50 & 4.58 & +2.92 & [-0.42, +7.08] \\
AIME25 & Pass@8 & 19.26 & 12.80 & +6.46 & [-3.00, +16.72] \\
OlympiadBench & Acc@8 & 30.84 & 31.57 & -0.72 & [-2.45, +0.95] \\
OlympiadBench & Pass@8 & 45.24 & 46.56 & -1.33 & [-3.87, +1.24] \\
\midrule
Macro average & Acc@8 & 28.01 & 27.76 & +0.25 & [-1.48, +2.01] \\
Macro average & Pass@8 & 41.85 & 40.53 & +1.31 & [-1.41, +4.15] \\
\bottomrule
\end{tabular}
\end{table}

Table~\ref{tab:app-full-fresh-ci} gives positive macro point estimates for
HISPO on both Acc@8 and Pass@8, but both macro intervals cross zero. MATH500 is
the clearest positive case, with positive intervals for both metrics. Minerva
and AIME24 are the main unstable cases, motivating a focused repeat diagnostic
on these two benchmarks (Table~\ref{tab:app-focused-minerva-aime24-ci}).

\begin{table}[t]
\caption{Focused repeat selected-checkpoint diagnostic on Minerva and AIME24.
Values are percentages; intervals are 95\% paired bootstrap intervals over
matched problem IDs.}
\label{tab:app-focused-minerva-aime24-ci}
\centering
\scriptsize
\setlength{\tabcolsep}{3.0pt}
\begin{tabular}{@{}p{0.20\linewidth}p{0.10\linewidth}rrrp{0.12\linewidth}@{}}
\toprule
Benchmark & Metric & HISPO & GSPO & $\Delta$ & 95\% CI \\
\midrule
Minerva & Acc@8 & 17.14 & 18.11 & -0.97 & [-2.90, +0.92] \\
Minerva & Pass@8 & 28.09 & 28.83 & -0.74 & [-3.93, +2.42] \\
AIME24 & Acc@8 & 7.50 & 5.00 & +2.50 & [-0.42, +5.83] \\
AIME24 & Pass@8 & 21.28 & 12.32 & +8.96 & [+1.09, +18.80] \\
\midrule
Minerva+AIME24 macro & Acc@8 & 12.32 & 11.55 & +0.77 & [-1.12, +2.64] \\
Minerva+AIME24 macro & Pass@8 & 24.68 & 20.58 & +4.11 & [-0.02, +9.05] \\
\bottomrule
\end{tabular}
\end{table}

Table~\ref{tab:app-focused-minerva-aime24-ci} shows that these small and
difficult benchmarks are sensitive to sampled completions. Minerva remains
slightly negative but no longer has a strictly negative Acc@8 interval, while
AIME24 has positive point estimates, with a strictly positive Pass@8 interval.
We therefore treat both reruns as diagnostics of sampling variability rather
than definitive per-benchmark statistical confirmation.

\subsection{Existing Assets and Licenses}
\label{app:asset-licenses}

Table~\ref{tab:appendix-asset-licenses} lists the existing assets used in the
experiments. We use these assets only for research training and evaluation, cite
their original sources, and do not redistribute benchmark contents or model
weights as part of this paper.

\begin{table}[t]
\caption{Existing assets used in the experiments.}
\label{tab:appendix-asset-licenses}
\centering
\scriptsize
\begin{tabular}{@{}p{0.30\linewidth}p{0.20\linewidth}p{0.35\linewidth}@{}}
\toprule
Asset and source & Use & License / terms \\
\midrule
Qwen3-1.7B-Base; \url{https://huggingface.co/Qwen/Qwen3-1.7B-Base}
& Base policy model
& Apache 2.0 \\

MATH-lighteval; \url{https://huggingface.co/datasets/DigitalLearningGmbH/MATH-lighteval}
& Training data
& MIT \\

MATH500; \url{https://huggingface.co/datasets/HuggingFaceH4/MATH-500}
& Evaluation benchmark
& Source split is from OpenAI
PRM800K / MATH splits, released under MIT \\

AMC23; \url{https://huggingface.co/datasets/math-ai/amc23}
& Evaluation benchmark
& Copyrighted by the Mathematical Association of America /
American Mathematics Competitions. Used only for evaluation; not redistributed \\

Minerva; \url{https://huggingface.co/datasets/math-ai/minervamath}
& Evaluation benchmark
& Primarily sourced from MIT OpenCourseWare, whose default terms
are CC BY-NC-SA 4.0. Used only for evaluation; not
redistributed \\

AIME24; \url{https://huggingface.co/datasets/HuggingFaceH4/aime_2024}
& Evaluation benchmark
& Copyrighted by the Mathematical Association of America. Used
only for evaluation; not redistributed \\

AIME25; \url{https://huggingface.co/datasets/math-ai/aime25}
& Evaluation benchmark
& Apache 2.0 on the Hugging Face dataset card \\

OlympiadBench; \url{https://huggingface.co/datasets/math-ai/olympiadbench}
& Evaluation benchmark
& Official OlympiadBench GitHub
repository is MIT licensed. Used only for evaluation; not redistributed \\

\texttt{verl}; \url{https://github.com/verl-project/verl}
& RL training platform
& Apache 2.0 \\

Hugging Face Transformers; \url{https://github.com/huggingface/transformers}
& Model implementation utilities
& Apache 2.0 \\

Hugging Face PEFT; \url{https://github.com/huggingface/peft}
& LoRA implementation utilities
& Apache 2.0 \\
\bottomrule
\end{tabular}
\end{table}

\subsection{Societal Impact}
\label{app:societal-impact}

HISPO may support beneficial applications such as STEM education, scientific
problem solving, and coding assistance by improving mathematical RLVR
post-training. It may also strengthen dual-use automation, produce more
convincing but incorrect reasoning traces, or encourage over-reliance in
high-stakes settings. We view HISPO as a research contribution to
policy-optimization granularity rather than a deployment-ready system; use in
consequential applications requires domain-specific evaluation and safeguards.

\clearpage

\section*{NeurIPS Paper Checklist}



\begin{enumerate}

\item {\bf Claims}
    \item[] Question: Do the main claims made in the abstract and introduction accurately reflect the paper's contributions and scope?
    \item[] Answer: \answerYes{}
    \item[] Justification: The abstract and introduction state HISPO's segment-level clipped importance-sampling objective and scope. The main empirical claims are supported by Table~\ref{tab:main-selected-checkpoint-at8} and caveated in Section~\ref{sec:limitations}.


\item {\bf Limitations}
    \item[] Question: Does the paper discuss the limitations of the work performed by the authors?
    \item[] Answer: \answerYes{}
    \item[] Justification: Section~\ref{sec:limitations} discusses causal uncertainty, limited scale, LoRA training, mathematical-RLVR scope, response length, clipping pressure, and future validation.


\item {\bf Theory assumptions and proofs}
    \item[] Question: For each theoretical result, does the paper provide the full set of assumptions and a complete (and correct) proof?
    \item[] Answer: \answerNA{}
    \item[] Justification: The paper does not present formal theorems. It provides objective definitions and gradient decompositions under stated simplifying assumptions.


    \item {\bf Experimental result reproducibility}
    \item[] Question: Does the paper fully disclose all the information needed to reproduce the main experimental results of the paper to the extent that it affects the main claims and/or conclusions of the paper (regardless of whether the code and data are provided or not)?
    \item[] Answer: \answerYes{}
    \item[] Justification: Section~\ref{sec:experiments} and Appendix~\ref{app:implementation-details} specify the model, data, benchmarks, methods, checkpoint selection, evaluation protocol, implementation stack, LoRA settings, and hyperparameters.


\item {\bf Open access to data and code}
    \item[] Question: Does the paper provide open access to the data and code, with sufficient instructions to faithfully reproduce the main experimental results, as described in supplemental material?
    \item[] Answer: \answerNo{}
    \item[] Justification: The submission includes an anonymized partial supplement with HISPO loss excerpts, configuration wrappers, selected evaluation/recompute scripts, result CSVs, figure assets, checksums, and provenance notes. It does not release full checkpoints, raw logs, full datasets, or a complete standalone reproduction package.


\item {\bf Experimental setting/details}
    \item[] Question: Does the paper specify all the training and test details (e.g., data splits, hyperparameters, how they were chosen, type of optimizer) necessary to understand the results?
    \item[] Answer: \answerYes{}
    \item[] Justification: Section~\ref{sec:experiments} and Appendix~\ref{app:implementation-details} specify the training setup, evaluation suite, metrics, compared methods, hyperparameters, decoding settings, and HISPO-specific parameters.


\item {\bf Experiment statistical significance}
    \item[] Question: Does the paper report error bars suitably and correctly defined or other appropriate information about the statistical significance of the experiments?
    \item[] Answer: \answerYes{}
    \item[] Justification: Appendix~\ref{app:uncertainty-diagnostics} reports paired bootstrap uncertainty diagnostics for fresh selected-checkpoint HISPO--GSPO evaluations. These intervals capture problem-level sampling variability, but not training-seed variance or checkpoint-selection uncertainty.


\item {\bf Experiments compute resources}
    \item[] Question: For each experiment, does the paper provide sufficient information on the computer resources (type of compute workers, memory, time of execution) needed to reproduce the experiments?
    \item[] Answer: \answerYes{}
    \item[] Justification: Appendix~\ref{app:compute-resources} reports GPU type and memory, wall-clock training/evaluation costs, checkpoint-sweep costs, total GPU-hours, and exclusion of preliminary debugging runs.

    
\item {\bf Code of ethics}
    \item[] Question: Does the research conducted in the paper conform, in every respect, with the NeurIPS Code of Ethics \url{https://neurips.cc/public/EthicsGuidelines}?
    \item[] Answer: \answerYes{}
    \item[] Justification: The work studies algorithmic RLVR optimization using public models and benchmarks, without private data, human-subject research, crowdsourcing, surveillance, or biometric data.


\item {\bf Broader impacts}
    \item[] Question: Does the paper discuss both potential positive societal impacts and negative societal impacts of the work performed?
    \item[] Answer: \answerYes{}
    \item[] Justification: Appendix~\ref{app:societal-impact} discusses potential benefits for STEM, science, and coding, as well as risks from dual-use automation, incorrect reasoning traces, and over-reliance.

    
\item {\bf Safeguards}
    \item[] Question: Does the paper describe safeguards that have been put in place for responsible release of data or models that have a high risk for misuse (e.g., pre-trained language models, image generators, or scraped datasets)?
    \item[] Answer: \answerNA{}
    \item[] Justification: The paper does not release a new pretrained model, high-risk dataset, or deployed system.


\item {\bf Licenses for existing assets}
    \item[] Question: Are the creators or original owners of assets (e.g., code, data, models), used in the paper, properly credited and are the license and terms of use explicitly mentioned and properly respected?
    \item[] Answer: \answerYes{}
    \item[] Justification: Appendix~\ref{app:asset-licenses} lists existing models, datasets, benchmarks, and software assets, including sources, uses, licenses, or source terms.


\item {\bf New assets}
    \item[] Question: Are new assets introduced in the paper well documented and is the documentation provided alongside the assets?
    \item[] Answer: \answerYes{}
    \item[] Justification: The submission includes an anonymized partial supplement with HISPO code excerpts, configs, scripts, result CSVs, figure assets, README, requirements, checksums, and provenance notes. It is documented as a partial review artifact rather than a full standalone release.


\item {\bf Crowdsourcing and research with human subjects}
    \item[] Question: For crowdsourcing experiments and research with human subjects, does the paper include the full text of instructions given to participants and screenshots, if applicable, as well as details about compensation (if any)? 
    \item[] Answer: \answerNA{}
    \item[] Justification: The paper does not involve crowdsourcing, human-subject experiments, user studies, or new human annotations.


\item {\bf Institutional review board (IRB) approvals or equivalent for research with human subjects}
    \item[] Question: Does the paper describe potential risks incurred by study participants, whether such risks were disclosed to the subjects, and whether Institutional Review Board (IRB) approvals (or an equivalent approval/review based on the requirements of your country or institution) were obtained?
    \item[] Answer: \answerNA{}
    \item[] Justification: The paper does not involve human-subject or crowdsourcing research, so IRB or equivalent approval is not applicable.


\item {\bf Declaration of LLM usage}
    \item[] Question: Does the paper describe the usage of LLMs if it is an important, original, or non-standard component of the core methods in this research? Note that if the LLM is used only for writing, editing, or formatting purposes and does \emph{not} impact the core methodology, scientific rigor, or originality of the research, declaration is not required.
    \item[] Answer: \answerYes{}
    \item[] Justification: The core experiments fine-tune and evaluate Qwen3-1.7B-Base, as described in Section~\ref{sec:experiments} and Appendix~\ref{app:implementation-details}. No external LLM is used for labeling, evaluation, or as a non-standard component of the method.


\end{enumerate}

\end{document}